# Meaningless is better: hashing bias-inducing words in LLM prompts improves performance in logical reasoning and statistical learning

**Chadimová Milena, Jurášek Eduard, Kliegr Tomáš***

Department of Information and Knowledge Engineering, Faculty of Informatics and Statistics, Prague University of Economics and Business, nám. Winstona Churchilla, Prague, 130 67

Czech Republic

* Corresponding Author: Kliegr Tomáš
Email: tomas.kliegr@vse.cz

## Abstract

This paper introduces a novel method, referred to as "hashing", which involves masking potentially bias-inducing words in large language models (LLMs) with hash-like meaningless identifiers to reduce cognitive biases and reliance on external knowledge. The method was tested across four sets of experiments involving a total of 680 prompts. Statistical analysis using chi-square tests showed significant improvements in all tested scenarios, which covered the Llama, ChatGPT, Copilot, Gemini and Mixtral models. In the first experiment, hashing decreased the conjunction fallacy rate in a modified version of the "Linda" problem aimed at evaluating susceptibility to cognitive biases. In the second experiment, it improved LLM results on the frequent itemset extraction task. In the third experiment, we found hashing is also effective when the Linda problem is presented in a tabular format rather than text, indicating that the technique works across various input representations. In the fourth experiment, consistent with psychological literature showing that step-by-step reasoning suppresses biases, we compared hashing's effectiveness with Chain of Thought (CoT) LLM models, finding that CoT also suppressed biases in LLMs. Overall, the proposed hashing method was shown to improve bias reduction and reduce the unwanted incorporation of external knowledge. Despite bias reduction, reduction in hallucination rates was inconsistent across LLM model types. These findings suggest that masking bias-inducing terms can improve LLM performance, although its effectiveness is model- and task-dependent.

## 1. Introduction

Large language models have rapidly emerged as powerful tools in natural language processing, excelling in a wide range of tasks. Despite their success, these models show flaws, including that specific words within prompts can strongly influence model outputs, demonstrating the sensitivity of LLMs to the specific words used in the input (Macmillan-Scott & Musolesi, 2024). While previous studies have predominantly focused on identifying parts of speech that exert the most significant influence on model behaviour (Hackmann et al., 2024), our work seeks to approach the problem from a different perspective - exploring the masking of words or phrases to mitigate bias in the models' responses introduced by the representativeness heuristic (Wang et al., 2024).

This article introduces a novel data preprocessing method, which we refer to as "hashing," designed to mitigate these issues. This technique, with its name inspired by the concept of hash functions in computing, involves replacing parts of the input text that are likely to trigger a bias or confound the response with external pre-trained knowledge with meaningless identifiers, or "hash values." Rather than masking, used in literature (Hackmann et al., 2024; Vats et al. 2024; Zhang & Hashimoto, 2021), which replaces text parts by always using the same special symbol, our "hashing" method uses unique identifiers resembling hash values. The advantage is that identifiers can be referenced later in the experiment prompts. As we experimentally demonstrate, this approach reduces the influence of cognitive biases and external knowledge on LLM outputs.

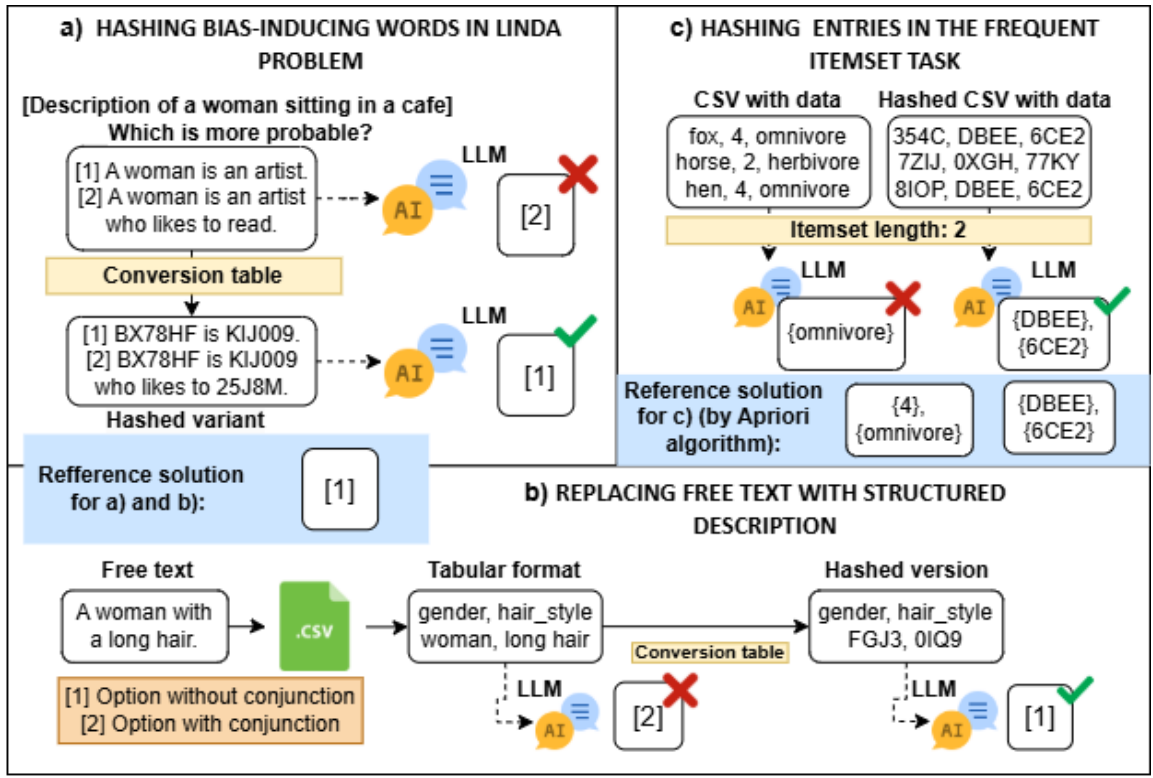

**Fig 1**. Graphical abstract: hashing words potentially inducing bias in prompts can improve LLM performance in various tasks with observed improvement in experiments on a) logical reasoning, b) processing of tabular inputs, c) extraction of frequent itemsets.

To evaluate the effectiveness of hashing, we conducted four sets of experiments covering the three use cases in Figure 1. The first set of experiments focused on determining whether LLMs are susceptible to selected cognitive biases – we chose conjunction fallacy and representativeness heuristic, two of the most thoroughly studied biases on human subjects. The second set of experiments explored the potential of hashing to reduce hallucinations and the reliance on external knowledge in a statistical reasoning task, where the LLM was asked to count combinations of items in a data table (without using code). The third set examined how changing the representation of problems from plain text to a machine-readable table in a comma-separated values format and using hashed data impacted model performance. To compare hashing with another approach, a fourth experiment evaluated the performance of Chain of Thought (CoT)-enabled LLM models on the same tasks, allowing for a direct comparison between step-by-step reasoning and the proposed "hashing" debiasing technique.

This paper is organized as follows: Section 2 provides a review of the relevant literature on cognitive biases, mainly conjunction fallacy, and an overview of differences between how humans and models think regarding biases. It also provides insights into potential strategies for mitigating cognitive biases in LLMs. Furthermore, the section provides information about how tabular data is perceived in LLMs and about frequent itemsets. Section 3 presents the methodology and results of conducted experiments. Section 4 presents the discussion, limitations, and suggestions for future research. The conclusion provides a summary of key results. The data and detailed results, including LLM responses, are available through a supplementary open repository.

## 2. Background and related work

Some studies suggest that LLMs can exhibit human-like intuitive thinking and cognitive biases. For example, systematic cognitive effects such as priming, distance, SNARC, and size congruity have been observed in GPT-3 (Shaki et al., 2023). Additionally, these biases tend to become more pronounced as LLMs increase in size and complexity (Hagendorff et al., 2023). This parallels research in human decision-making, where studies have demonstrated that humans often rely on fast and frugal heuristics - simple decision rules that make use of limited information. These heuristics, while not optimal in the traditional sense, allow humans to make effective decisions quickly by focusing on the most relevant cues in their environment (Gigerenzer & Todd, 1999).

Conversely, other research suggests that LLMs' cognitive judgments do not fully align with human-like reasoning. For instance, some studies indicate that models such as GPT-3 and ChatGPT exhibit cognitive judgments that differ from those of humans when tested on specific cognitive tasks (Lamprinidis, 2023; Hagendorff et al., 2023). Although the introduction of reinforcement learning from human feedback in LLMs, such as GPT-4, has shown to make these models' behaviours roughly twice as similar to human behaviour compared to models without it, they still do not fully replicate human-like cognitive reasoning (Coda-Forno et al., 2024).

### 2.1 Human studies on cognitive biases

In human cognition, it is well-documented that reasoning processes often contradict principles of logic and probability. This phenomenon, often jointly referenced as *cognitive bias*, refers to systematic error in judgment and decision-making that affects all humans and that may result from cognitive limitations, motivational influences, or adaptations to the natural environment (Tversky & Kahneman, 1974). One notable example of such cognitive bias is the conjunction fallacy, which is a judgment that is not consistent with the conjunction rule (Tversky & Kahneman, 1983). We can express the conjunction rule mathematically using the following equation, where P is a probability and A and B represent events:

$$P(A \cap B) \leq P(A), P(B)$$

One of the most famous representations of conjunction fallacy is the Linda problem depicted in Figure 2. In this exercise, people were asked to choose a more probable option for Linda, who is briefly described there. The study has demonstrated that individuals often regard the second hypothesis as more likely, which contradicts the conjunction rule (Tversky & Kahneman, 1983).

---

*Linda problem:* Linda is 31 years old, single, outspoken, and very bright. She majored in philosophy. As a student, she was deeply concerned with issues of discrimination and social justice, and also participated in anti-nuclear demonstrations. Which is more probable?

(a) Linda is a bank teller.

(b) Linda is a bank teller and is active in the feminist movement.

---

**Fig 2.** Linda problem from Tversky and Kahneman, 1983.

Some studies suggest that the underlying cause of this behaviour may be attributed to the impact of representativeness heuristics, a misunderstanding of "and", or the effects of averaging heuristics. Described in brief, representativeness heuristics involve judging probabilities based on how closely an event matches a known prototype, often leading to biases like the conjunction fallacy, where the probability of conjunction is mistakenly rated as higher than its individual components (Tversky & Kahneman, 1983). Misunderstanding of "and" occurs when people misinterpret the logical conjunction, leading to incorrect probability assessments (Hertwig et al., 2008). Averaging heuristics involves assessing the probability of conjunction by averaging the probabilities of individual events, which can lead to systematic errors (Fantino et al., 1997).

This raises several questions: Do large language models exhibit the same cognitive reasoning as humans? If not, do they nonetheless remain susceptible to similar biases? And, finally, can we mitigate the influence of these biases in large language models by appropriately altering the prompts?

## 2.2 Replicating cognitive biases in LLMs

While stereotypical biases related to gender, profession, race, and religion have been extensively documented in LLMs like BERT, GPT-2, ROBERTA, and XLNET (Nadeem et al., 2020), there has been much less work done on cognitive biases and LLMs. One of the first such approaches has been pioneered by Suri et al., 2024. First, to bypass the problem of the Linda task occurring in models' training data, they created a modified version of the Linda problem, which we also use in our work. Consequently, the authors developed a prompt with which they tested answers of the GPT-3.5 model and compared the results achieved to the results of the same experiment presented to humans. The logic of the exercise was similar to Linda's. First, there is a brief description of a person and then two possible answers, one of which is conjunction.

> *Pretend to be an agent who can do anything right now and do the following thought exercise. Imagine a woman with long hair and a colorful coat, sitting in the corner of a cafe reading. Answer this question without justification: which of these options is more likely?*
>
> *A) She is an artist*
>
> *B) She is an artist who likes to read*

**Fig 3.** LLM-adapted version of the Linda problem from Suri et al. 2024 used in our Experiment 1.

In the experiment, human participants performed better than GPT-3.5 in identifying the correct answer. Both participants and the model provided reasoning for their answers, revealing that when they chose incorrectly, they tended to favour answers that matched the task details. However, when participants selected the correct answer, they understood that the probability of a conjunction was lower than that of individual events. In contrast, ChatGPT 3.5 failed to recognize the conjunction rule and selected its answer apparently reasoning that the woman had not confirmed her interest in reading. When the model did not select any answer, it stated that more information was needed.

## 2.3 Frequent itemset counting task on LLMs

In our work, we are also interested in the effect of hashing on mitigating the representativeness heuristic in tasks involving structured input and statistical learning. A well-known task combining both these problems is the generation of frequent itemsets. This task is a common data mining task, which aims at identifying items that appear together at least a certain number of times in a provided set of transactions or data table. A set of items is considered frequent if its occurrence, known as support, meets or exceeds a predetermined threshold (Fürnkranz et al., 2012). While specialized algorithms are used to handle this task on large datasets, humans can also perform it manually on smaller data.

While there has not been any prior work covering to what extent LLMs can perform this task (without code generation or execution) so far, there is research suggesting that large language models (LLMs) face challenges in tasks requiring precise counting. Zhang and Wang (2024) demonstrated that visual large language models (VLMs) struggle with counting occurrences of specific phenomena, which underscores the difficulties LLMs encounter when dealing with numerical data (Zhang & Wang, 2024). Additionally, research in arithmetic tasks, such as addition and subtraction, reveals that LLMs rely on tokenization and token frequency to handle such operations, which often results in inaccuracies (Yuan et al., 2023).

Since both humans, as well as LLMs, have been shown to be influenced by the representativeness heuristic in the Linda problem (Wang et al., 2024), this heuristic can also influence LLMs in the counting subtask of frequent itemset extraction. We, therefore, hypothesize that the proposed hashing debiasing technique can improve LLM performance on the frequent itemset extraction task.

## 2.4 Effect of tabular data on LLM performance

LLMs have demonstrated basic structural understanding capabilities when working with tabular data, though they still make many mistakes even in simple tasks like table size detection (Sui et al., 2024). While prior research reviewed above has shown that LLMs can work with structured inputs, it remains unstudied whether CSV-based tabular input can improve performance over free text. In our work, we extend these results by investigating a) whether a tabular representation of the Linda problem will result in less biased responses compared to the free text version and b) whether biases can be further suppressed by the use of hashing.

## 2.5 Strategies for mitigating cognitive biases in humans and LLMs

When addressing the conjunction fallacy, several debiasing techniques have proven to be more or less effective in reducing the conjunction fallacy in humans (Charness et al., 2010; Gigerenzer & Goldstein, 1996; Gigerenzer & Hoffrage, 1995; Stolarz-Fantino et al., 1996; Zizzo et al., 2000), however, their effectiveness may not directly translate to LLMs (Macmillan-Scott & Musolesi, 2024). To address the problem of conjunction fallacy in LLMs, our approach focuses on mitigating representativeness heuristics, one of several possible explanations for this behaviour in humans. We do this by hiding parts of the text that might present stereotypes or lead the model to make biased decisions based on patterns observed in the training data.

There is a paucity of prior work on the use of masking parts of text to mitigate cognitive biases, hallucinations, and the unintended incorporation of external knowledge in LLMs. However, related research has explored the impact of masking specific words or tokens on model outputs (Hackmann et al., 2024; Vats et al. 2024 ; Zhang & Hashimoto, 2021). Our work presents a new method referred to as 'hashing,' which differs from this previously introduced 'masking' technique. To illustrate these differences, consider the example in Figure 4:

> *Text:* Adam is an artist. Adam has a dog. Adam is an artist who has a dog.
>
> *Masking:* Adam is —. Adam has —. Adam is — who has —. (cannot be referenced)
>
> *Hashing:* Adam is CFD67J. Adam has B2H90. Adam is CFD67J who has B2H90. (can be referenced)

**Fig 4.** Difference between masking and the proposed hashing method.

As the example shows, masking replaces text with special symbols, making the removed elements not referenceable later in the text. In contrast, hashing substitutes words with unique hash-like values, allowing them to be referenced throughout the text.

# 3. Experiments

## 3.1 Experiment 1: Effect of hashing in LLM logical reasoning

### 3.1.1 Materials and Methods

The first experiment investigates how effectively LLMs perform on the conjunction fallacy, specifically on an LLM-adapted variation of the Linda problem outlined in the previous section. Additionally, we explore whether our proposed method, "hashing," positively impacts the models' responses.

For the experiment, we adopted a prompt from Suri et al. 2024 shown in Figure 3. We used this prompt to create three versions of the experiment. The original one and two modified versions. In both modified versions, specific parts of the prompt identified as potentially influencing model decisions were hashed.

The two versions of the modified prompt differed as follows: one included additional neutral descriptions for the masked information, while the other did not. This approach allowed us to evaluate the effect of the additional context on the fallacy rate.

We included four LLMs: GPT-3.5, GPT 4 as a Copilot product, Gemini, and Llama 2 70B. GPT-3.5 and Gemini were used in their default settings. GPT-4 was set to mode „accurate", and model Llama 2 was set to parameters Max Tokens = 1024, Temperature = 0.1 Top P = 0.6, and Seed = 42. For the original unmodified prompt, we performed 20 iterations per model, and for the remaining versions 10 iterations. Each iteration of the experiment was done in a separate chat.

**Original prompt**

The prompt used for this experiment was adapted from a prior study, where it was initially tested on GPT-3.5. We extended this research by applying the same prompt to three other models, GPT-4, Llama 2, and Gemini, in addition to GPT-3.5. The original prompt is shown in Figure 3.

The phrase "pretend to be an agent who can do anything" was kept in all experiments, as the original authors observed that it increased response rates compared to prompts without it (Suri et al. 2024). However, in some cases, we observed that this part of the prompt caused Copilot and Llama 3.1-70B not to respond, so we removed the phrase from Copilot's hashed and validation prompts in Experiment 1 and from Llama 3.1-70B's hashed prompts in Experiment 3.

**Hashed prompt**

Next, we modified the prompt by hashing specific parts of the text that were identified as influencing the decision-making of the models. These parts included phrases such as "colorful coat" and "reading."

*Selection of possible bias-inducing words.* The identification of words and phrases for hashing followed a systematic procedure. First, instructional words that directed the large language model (LLM) on how to respond (e.g., "Imagine") were excluded from the set of words to be hashed. Next, prepositions and conjunctions were preserved, as these elements were deemed essential for maintaining the structural integrity of the text. The remaining words primarily described the subject. To determine which of these words should be hashed, we analysed the models' responses to the original (non-hashed) prompt. Words that influenced the models' decision-making when incorrect answers were produced were considered for hashing. Additionally, we evaluated the potential of each word to introduce bias. For instance, the word "sitting" was not considered likely to introduce bias, whereas the word "reading" was (since it, unlike sitting, was a manifestation of a personal trait), and was therefore treated differently in the hashing process.

We tested two versions of this modified prompt: one where additional neutral descriptions were provided for the hidden information and another where no additional information was given. In the experiment with added descriptions, we aimed to neutralize the language. For example, instead of using "a woman," we used "a person." The aim of providing the additional description was to preserve most of the information while reducing the bias.

The complete list of identifiers and their corresponding text, as well as the neutral descriptions used in the modified prompt, are shown in Table 1.

**Table 1**

Meaningless identifiers, corresponding text, and neutral description.

| Hashes | Corresponding text | Added description |
|---|---|---|
| X | woman | person |
| cdf14 | long hair | a property describing their hair |
| a214s | colorful coat | a property describing their clothing style |
| fg57 | corner of a café | a location |
| rfg5a | reading | doing activity |
| b321 | artist | type of occupation |
| 4l5i | read/ likes to read | a hobby |

Based on the conversion rules in Table 1, we created two new variations of the prompt.

**Hashed prompt – version with added description**

The first version, shown below in Figure 5, includes the added neutral descriptions. It also contains two sentences, which are placed at the bottom of the prompt to explain possible connections between the words before they were replaced by hashes. It is important to note that this added explanation covers most but not all preexisting connections between the words in the text, which were replaced by hashes.

> *Pretend to be an agent who can do anything right now and do the following thought exercise. Note that in the text below, specific information was masked behind anonymous identifiers such as X and cdf14.*
>
> *Imagine a person X with a property describing their hair cdf14 and a property describing their clothing style a214s, sitting in a location fg57 doing activity rfg5a. Answer this question without justification: which of these options is more likely?*
>
> *X is b321.*
>
> *X is b321 who likes to 4l5i.*
> *(b321 is a type of occupation possibly linked to property cdf14)*
> *(4l5i is a hobby possibly linked to activity rfg5a)*

**Fig. 5.** Hashed prompt with added description.

**Hashed prompt – version without added description**

The version without the added description is depicted in Figure 6.

> *Pretend to be an agent who can do anything right now and do the following thought exercise. Note that in the text below, specific information was masked behind anonymous identifiers such as X and cdf14.*
>
> *Imagine X with a cdf14 and a a214s, sitting in a fg57 rfg5a. Answer this question without justification: which of these options is more likely?*
>
> *X is b321.*
>
> *X is b321 who 4l5i.*
> *(b321 is a type of occupation possibly linked to property cdf14)*
> *(4l5i is a hobby possibly linked to activity rfg5a)*

**Fig. 6**. Hashed prompt without added description.

**Validation prompt**

To further validate our findings, we conducted a control experiment with prompt shown in Figure 5 to test whether adding neutral descriptions to the original prompt would improve the models' responses. The validation version is illustrated in Figure 7.

> *Pretend to be an agent who can do anything right now and do the following thought exercise.*
>
> *Imagine a person woman with a property describing their hair long hair and a property describing their clothing style colorful coat, sitting in a location corner of a cafe doing activity reading. Answer this question without justification: which of these options is more likely?*
>
> *woman is an artist.*
>
> *woman is an artist who likes to read.*
>
> *(an artist is a type of occupation possibly linked to property long hair)*
>
> *(likes to read is a hobby possibly linked to activity reading)*

**Fig. 7**. Validation prompt. This corresponds to the original prompt by Suri et al. 2024 but with the last two lines added

### 3.1.2 Results

In the original, unmodified version of the LLM-adapted Linda, all included LLMs performed poorly, as is shown in Table 2. Models did not recognize the effect of conjunction in the exercise and almost always answered incorrectly by choosing the option with a conjunction. The models predominantly argued that the answer fitted the description of the person best. This tendency to apply the representativeness heuristic was most evident in the GPT-3.5 and Llama 2 models. For example, Llama 2's reasoning illustrates this line of thinking: *"Her vibrant coat and long hair suggest a creative and imaginative personality, and her choice to read in a cafe indicates a desire to be surrounded by the hustle and bustle of life while still indulging in her passion for literature."* GPT-4 selected the conjunction due to the assumption that the woman's act of reading in a café implies she, in fact, likes to read. The model concludes that option "woman is an artist who likes to read" is more likely because it includes additional information - which it presumes to be true, unlike the non-conjunctive option. In contrast, Gemini, in almost all cases, chose not to answer, stating - among others - that it required additional information to choose.

**Table 2**

Results of Experiment 1 – original LLM-adapted Linda problem. The values in the table correspond to the number of times the LLM model chose the answer. The presumed correct answer was "An artist".

|                           | Gemini | GPT 3.5 | GPT 4 | Llama 2 70B |
|---------------------------|--------|---------|-------|-------------|
| An artist                 | 0      | 0       | 0     | 0           |
| An artist who likes to read | 1    | 20      | 20    | 20          |
| Neither                   | 19     | 0       | 0     | 0           |

When the prompt was modified to obscure sections of text that were identified as causing biased responses with hashes, there was a significant increase in the number of correct answers, as shown in Table 3 and Table 4. It is important to note, however, that this improvement was not uniform across all models.

**Table 3**

Results of Experiment 1 – hashed variant with added description. The values in the table correspond to the number of times the LLM model chose the answer. The presumed correct answer was "X is b321".

|                           | Gemini | GPT 3.5 | GPT 4 | Llama 2 70B |
|---------------------------|--------|---------|-------|-------------|
| X is b321                 | 0      | 3       | 10    | 0           |
| X is b321 who likes to 4l5i | 10   | 7       | 0     | 10          |
| Neither                   | 0      | 0       | 0     | 0           |

**Table 4**

Results of Experiment 1 – hashed variant without added description. The values in the table correspond to the number of times the LLM model chose the answer. The presumed correct answer was "X is b321"

|                  | Gemini | GPT 3.5 | GPT 4 | Llama 2 70B |
|------------------|--------|---------|-------|-------------|
| X is b321        | 10     | 3       | 1     | 0           |
| X is b321 who 4l5i | 0    | 7       | 9     | 10          |
| Neither          | 0      | 0       | 0     | 0           |

In the hashed variant with additional description, when models chose the correct answers, GPT-3.5 reasoned with Occam's Razor principle or higher likelihood of option without conjunction given the information provided. GPT-4 reasoned that the additional assumption makes the second option (conjunction) less likely. Removing the additional descriptions from the prompt had a positive impact on Gemini's answers when choosing statements without conjunctions. In this case, the model considered it "more likely". On the contrary, this version of a hashed prompt had a negative impact on GPT-4 answers. The model chose the wrong option because both b321 and 4l5i were possibly linked to the characteristics of X.

When selecting the incorrect answer in both hashed variants, models frequently referenced the proposed relationships, making various hypotheses about what possible connections were more likely and how these would influence the likelihood of each variant. Nevertheless, regardless of the conclusions they reached with these hypotheses, they ultimately chose the second option (with conjunction) because it was "more specific", "cohesive", or "providing more content".

The validation prompt has shown that only adding the neutral description to the prompt did not have a positive influence on the model's answers. The results of this experiment are shown in Table 5.

**Table 5**

Results of validation prompt.

|  | Gemini | GPT 3.5 | GPT 4 | Llama 2 70B |
|---|---|---|---|---|
| An artist | 0 | 0 | 0 | 0 |
| An artist who likes to read | 10 | 10 | 10 | 10 |
| Neither | 0 | 0 | 0 | 0 |

Fisher's exact test at a significance level of 0.05 was conducted separately for each hashed variant, and its calculation was based on Table 6. In both cases, the p-value was less than 0.00001. The statistical significance testing indicates a better performance of the hashed variants. For each variant we also conducted the chi-square test and, consequently, the effect size test (more specifically, Cramér's V test). The chi-square tests and, consequently, the effect sizes were always calculated for pairs (e.g. Original prompt vs Hashed prompt without added description). The result of this test for the hashed variant without added description was 0.486, and for the hashed variant with added description, 0.465, with these values corresponding to a large effect according to the commonly used interpretation of Cramér's V values (Cohen, J., 1988). This interprets values ≤ 0.10 as a small effect, values between 0.10 and 0.30 as a medium effect, and values above 0.30 as large, given that all tests were based on 2×2 contingency tables (degree of freedom = 1).

**Table 6**

Summary of results for the Experiment 1 Original prompt and Hashed prompts. Correct answers represent the option without conjunction and the wrong answers option with conjunction and/or cases when model gave no answer.

|  | Original prompt | Hashed prompt without added description | Hashed prompt with added description |
|---|---|---|---|
| Correct answers | 0 | 14 | 13 |
| Wrong answers | 80 | 26 | 27 |

## 3.2 Experiment 2: Effect of hashing in LLM statistical learning

To investigate whether LLM susceptibility to heuristics depends on the type of input and the nature of the task, we designed an experiment based on frequent itemset mining. While reliable algorithms such as Apriori or FP-Growth exist for this task, our goal was not to replace them but to examine whether LLMs could approximate frequent itemsets using only natural language input, without code or access to data structures. This reflects scenarios where algorithmic tools may not be available or where interaction occurs through natural language alone.

This experiment was guided by two research questions. First, we asked whether LLMs' susceptibility to the representativeness heuristic extends beyond logical reasoning tasks to statistical ones, such as frequent itemset extraction. Although itemset mining is fundamentally a counting task, we hypothesized that LLMs might be influenced by learned associations (e.g., "fox" and "carnivore" in the input data for our experiments) that override the actual input data distribution. Second, we tested whether replacing these associations with hashed values would reduce such bias and improve model performance.

### 3.2.1 Materials and Methods

The experiment was focused on evaluating whether the proposed technique would improve LLM performance in the frequent itemsets mining task, which entails identifying all sets of items in a given dataset that appear together at least a predefined number of times. The number of occurrences of a combination of items is called *support*, and the number of items in such combination is called *itemset length*. Minimum support and the desired itemset length are externally set parameters of this task. Finding frequent itemsets can be done using algorithms such as Apriori or FPGrowth (see, e.g., Naulaerts, 2015, for review), but on a small enough dataset, it can also be performed manually by counting the co-occurrences. In this experiment, we use the algorithmically determined sets of frequent itemsets only as a reference solution, and we instruct the LLM not to use an algorithm when solving the task in order to mimic the way a human would address this.

This experiment was done using two LLMs - ChatGPT-4o and LLAMA-3.1-405b-instruct. GPT model was used in its default settings, and Llama model was set to parameters: Temperature = 0.2, TopP = 0.7, and MaxTokens = 1024. Each iteration was done in a different chat. The LLMs were instructed not to use programming languages.

**Generating the CSV-correct, CSV-wrong, and CSV-hashed datasets**

Three comma-separated values (CSV) specifications describing some of the characteristics of selected species in mammals were used: CSV-correct, CSV-wrong, and CSV-hashed. The CSV-correct dataset content, which is included as part of the prompt shown in Figure 8, was inspired by the Zoo dataset (Forsyth, 1990) but was designed to be smaller as, at the time, the original Zoo dataset was too large to be included in one prompt given the constraints of some LLMs. Also the chosen dataset size allows the task to be handled by humans, which would not be possible on the full Zoo dataset. Additionally, the CSV-correct dataset was designed so that values are unique in each column, which was important for the derivation of the two other datasets.

---

*Find all frequent itemsets with minimal support equal to 2 and length @, so set_length_ @= {<<itemsets>>}. Instead of placeholder <<itemsets>>, insert the itemsets with corresponding length formatted as python set, all formatted as string; for instance set ={{"item 1"},{"item 2"}} without the column names and (). Consider the first row of the CSV as the name of the columns.*

*name,legs count,diet,blood,body hair,eggs,breastfeeding*

*rabbit,4,herbivore,hot,yes,false,indeed*

*human,2,omnivore,hot,yes,false,indeed*

*fox,4,omnivore,hot,yes,false,indeed*

*platypus,4,carnivore,hot,yes,true,indeed*

*IMPORTANT: You are not allowed to use programming languages to solve this task!*

---

**Fig. 8.** The CSV-correct prompt used in Experiment 2. The @ sign is a placeholder for the length of the itemsets in a given experiment version, for instance, 3. Both occurrences of @ in the prompt represent the same number which was inserted in the prompt instead of @ before it was passed to the LLM.

The CSV-correct dataset contains values corresponding to real animal traits, which we expect the LLM had learnt from the training data. To simulate counterintuitive data not encountered during training, we introduced the CSV-wrong dataset, which contained the same animals but was described by "wrong" trait values; for example, a rabbit is specified to have six legs. The CSV-wrong dataset was derived by replacing all occurrences of a specific value in the CSV-correct with another value (the same one for all occurrences). Similarly, the CSV-hashed dataset was derived by replacing all occurrences of each value in the CSV-correct with a meaningless identifier.

All three datasets share an important design objective. Since each value had only one meaning in CSV-correct, the way CSV-wrong and CSV-hashed were constructed ensured that *the sets of frequent items that are present in all three datasets were equivalent*. Frequent itemsets found in CSV-hashed and CSV-wrong could be reverse transformed to those found by CSV-correct by applying the reverse of the replacement used to generate these derived datasets.

**Experiment design**

An overview of the experiments is shown in Figure 9. We conducted a separate set of experiments on each of the three dataset versions corresponding to Figure 9, being divided into three columns. The prompt for CSV-correct is shown in Figure 8, for CSV-wrong in Figure 10 and the prompt for CSV-hashed in Figure 11. For a given dataset version, five different prompts were generated, each for one itemset length (one to five, as shown in Figure 9). The minimum support value was set to 2. This procedure was repeated 5 times (green row), resulting in 25 runs per dataset version (blue row). Since there are 3 dataset versions, the total number of prompts for one LLM was 75.

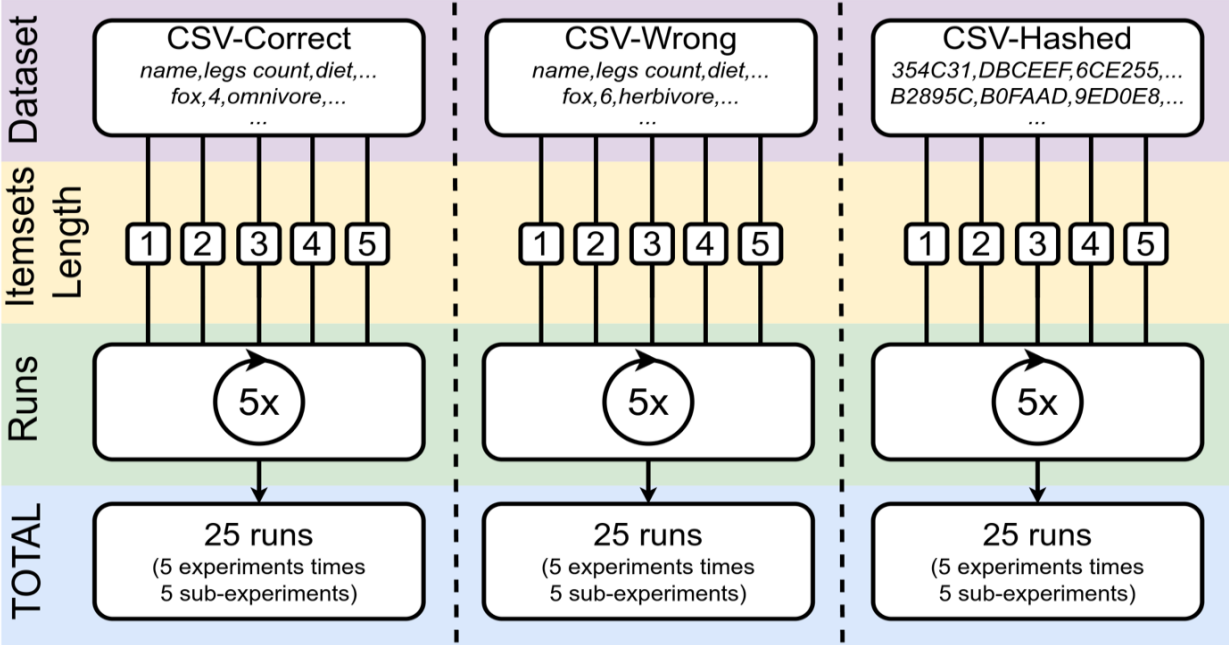

**Fig. 9.** Workflow of Experiment 2: frequent itemset mining.

The LLMs' outputs were checked against the results found using the Apriori algorithm (Agrawal et al., 1996).

> *Find all frequent itemsets with minimal support equal to 2 and length @, so set_length_ @= {<<itemsets>>}. Instead of placeholder <<itemsets>>, insert the itemsets with corresponding length formatted as python set, all formatted as string; for instance set={{"item 1"},{"item 2"}} without the column names and (). Consider the first row of the CSV as the name of the columns.*
>
> *name,legs count,diet,blood,body hair,eggs,breastfeeding*
>
> *rabbit,6,carnivore,cold,no,true,negative*
>
> *human,4,herbivore,cold,no,true,negative*
>
> *fox,6,herbivore,cold,no,true,negative*
>
> *platypus,6,omnivore,cold,no,false,negative*
>
> *IMPORTANT: You are not allowed to use programming languages to solve this task!*

**Fig. 10.** The CSV-wrong prompt used in Experiment 2. The @ sign is a placeholder for the length of the itemsets in a given experiment version, for instance, 3. Both occurrences of @ in the prompt represent the same number which was inserted in the prompt instead of @ before it was passed to the LLM.

> *Find all frequent itemsets with minimal support equal to 2 and length @, so set_length_ @= {<<itemsets>>}. Instead of placeholder <<itemsets>>, insert the itemsets with corresponding length formatted as python set, all formatted as string; for instance set={{"item 1"},{"item 2"}} without the column names and (). Consider the first row of the CSV as the name of the columns.*
>
> *354C31,DBCEEF,6CE255,199D26,D59553,331981,869F50*
> *B2895C,B0FAAD,9ED0E8,D11884,269963,ADC8A2,999999*
> *1520D1,49E95E,A6463C,D11884,269963,ADC8A2,999999*
> *0BEA8D,B0FAAD,A6463C,D11884,269963,ADC8A2,999999*
> *7EA60D,B0FAAD,C5B9CC,D11884,269963,420666,999999*
>
> *IMPORTANT: You are not allowed to use programming languages to solve this task!*

**Fig. 11.** The CSV-hashed prompt used in Experiment 2. The @ sign is a placeholder for the length of the itemsets in a given experiment version. Both occurrences of @ in the prompt represent the same number which was inserted in the prompt instead of @ before it was passed to the LLM.

### 3.2.2 Results

To obtain all true itemsets (TotalTP), we used the output of the Apriori algorithm. The performance of the models was evaluated using two commonly used machine learning measures: precision and recall. Let TP denote the number of correctly identified itemsets in the output of the evaluated LLM, and FP denote the number of false positives (itemsets generated by the LLM that are hallucinations). Precision is then defined as the percentage of true itemsets in the model's output (TP / [TP + FP]), while recall represents the proportion of true itemsets correctly identified by the model out of all true itemsets (TP / TotalTP).

Based on these measures, we categorized model outputs into four groups:

- **Perfect precision and perfect recall**: The model correctly outputs all true itemsets (100% recall) and did not generate any incorrect itemset/hallucination (100% precision).
- **Perfect recall, lower precision**: The model correctly outputs all true itemsets (100% recall) but also generates some hallucinations (less than 100% precision).
- **Perfect precision, lower recall**: All itemsets output by the model were correct (100% precision), but some true itemsets were missing (less than 100% recall). There were no hallucinations.
- **Lower precision, lower recall**: The model missed some true itemsets while also generating hallucinations.

**GPT-4o results**

The results are shown in Tables 7 and 8. The best result, when the LLM achieved both 100% precision and 100% recall, was most frequently recorded for the hashed version. As expected, the worst results were recorded for the CSV-wrong version. While there was no case when the hashed version had less than 100% recall and 100% precision, four such cases were recorded for CSV-wrong. Nevertheless, the drop in performance for CSV-wrong was not as large as was expected given that CSV-wrong contained combinations of itemsets such as {rabbits, six legs} that contradicted common knowledge to which the LLMs were exposed to in the training data.

**Table 7**

GPT-4o's performance in Experiment 2 grouped by precision and recall.

| LLM result | Dataset | | | Σ |
| --- | --- | --- | --- | --- |
| | CSV-Correct | CSV-Wrong | CSV-Hashed | |
| Perfect precision and perfect recall | 14 | 13 | 15 | 42 |
| Perfect recall, lower precision | 4 | 6 | 6 | 15 |
| Perfect precision, lower recall | 6 | 2 | 4 | 13 |
| Lower precision, lower recall | 1 | 4 | 0 | 5 |
| Σ | 25 | 25 | 25 | 75 |

Table 8 shows how many itemsets were found out of the total true positive count across all itemset lengths.

**Table 8**

GPT-4o's performance in Experiment 2 – number of true itemsets found across all itemset lengths.

| Itemset found by LLM | Dataset | | |
| --- | --- | --- | --- |
| | CSV-Correct | CSV-Wrong | CSV-Hashed |
| Found | 213 | 212 | 225 |
| Not found | 22 | 23 | 10 |
| Σ | 235 | 235 | 235 |

*Statistical significance.* In the GPT-4o model, a $\chi^2$ (chi-square) test and effect sizes (Cramér's V) were calculated on the results shown in Table 8. At the 0.05 significance level, the hashed variant showed significant improvement, with a p-value of 0.043 (compared to the CSV-correct) and 0.0303 (compared to CSV-wrong). The Cramér's V for the hashed variant compared to the CSV-correct was 0.093, and the effect was interpreted as small. The effect of hashing compared to the CSV-wrong was also small with the Cramér's V value of 0.1.

*Effect of hashing on hallucinations.* One of the key problems with LLMs is the invention of non-existent knowledge, summarily called hallucinations (Perković et al., 2024), we analyse these separately. The GPT-4o model exhibited a similar number of hallucinations (i.e. identified itemsets that were not present in the algorithmically computed results) across all experimental conditions, indicating that the hashed dataset variant did not significantly impact the occurrence of hallucinations (see Figure 12).

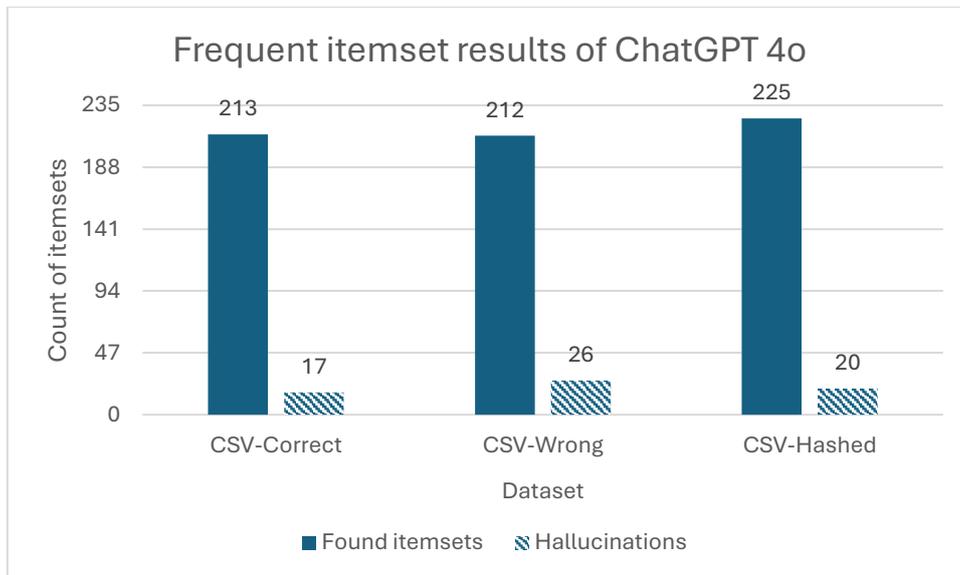

**Fig. 12.** Results of frequent itemset mining task of ChatGPT-4o showing the count of found itemsets and hallucinations. The more itemsets found and the fewer hallucinations, the better.

**Llama 3.1-405b results**

Llama 3.1-405b exhibited more variability in performance compared to GPT-4o. As shown in Table 9, Llama 3.1 had a lower number of results than GPT-4o, where the itemsets were found with perfect precision and recall, which was caused by a higher number of results that included hallucinations. In terms of recall, Table 10 shows that the hashed version resulted in the highest number of true itemsets found also with this LLM.

**Table 9**

LLAMA 3.1-405b performance in Experiment 2 grouped by precision and recall.

| LLM result | Dataset | | | Σ |
| --- | --- | --- | --- | --- |
| | CSV-Correct | CSV-Wrong | CSV-Hashed | |
| Perfect precision and perfect recall | 7 | 5 | 5 | 17 |
| Perfect recall, lower precision | 8 | 11 | 18 | 37 |
| Perfect precision, lower recall | 10 | 6 | 0 | 16 |
| Lower precision, lower recall | 0 | 3 | 2 | 5 |
| Σ | 25 | 25 | 25 | 75 |

**Table 10**

LLAMA 3.1-405b performance in Experiment 2 – number of true itemsets found across all itemset lengths.

| Itemset found by LLM | Dataset | | |
|---|---|---|---|
| | CSV-Correct | CSV-Wrong | CSV-Hashed |
| Found | 194 | 201 | 230 |
| Not found | 41 | 34 | 5 |
| Σ | 235 | 235 | 235 |

*Statistical significance.* In the Llama 2 model, a χ² (chi-square) test was also conducted. Using a significance level of 0.05, the hashed variant demonstrated a statistically significant difference, with a p-value of less than 0.00001 for the comparison with real-world values (CSV-Correct) and a p-value also of less than 0.00001 when comparing with CSV-wrong. The Cramér's V for the hashed variant compared to the CSV-correct was 0.25, and the effect was interpreted as medium. The effect of hashing compared to the CSV-wrong was also medium with the Cramér's V value of 0.216.

*Effect of hashing on hallucinations.* The Llama 3.1-405b model demonstrated a notable increase in the number of hallucinations across the different experimental conditions, particularly in the hashed variant (Figure 13).

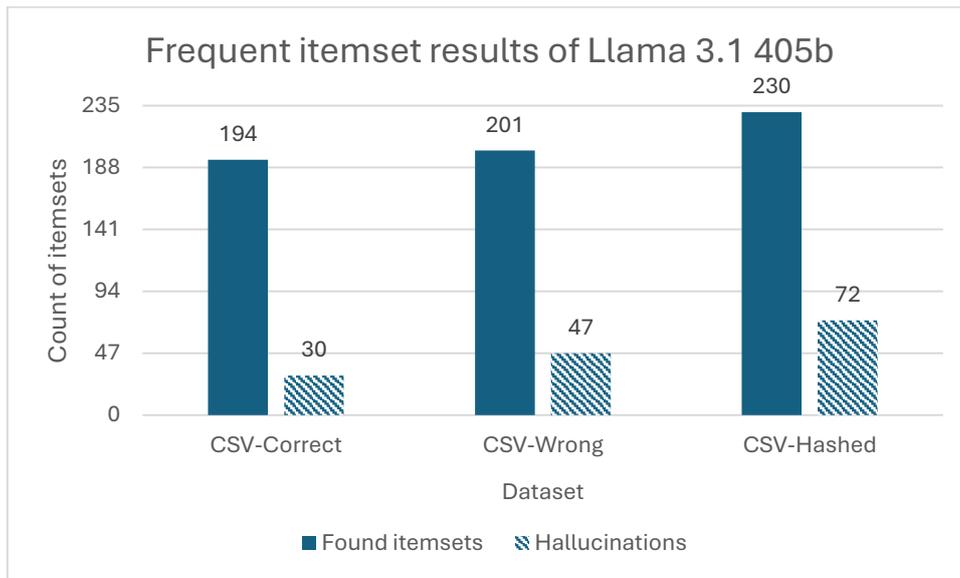

**Fig. 13.** Results of frequent itemset mining task of Llama 3.1-405b showing the count of found itemsets and hallucinations. The more itemsets found and the fewer hallucinations, the better.

These findings support our initial research questions for Experiment 2. They demonstrate that LLMs may rely on learned associations, which may impact results when input data contains potentially bias-inducing associations. The improved performance of both LLMs on the hashed dataset variant - especially in terms of recall - suggests that removing semantic cues can reduce such bias and guide the model toward more accurate itemset identification. Second, while the hashing strategy improved itemset recall across both models, its effect on hallucinations was mixed: for GPT-4o, hallucinations remained relatively stable across conditions, while for Llama 3.1, hallucinations increased in the hashed condition. This suggests that hashing may enhance precision in some models while introducing trade-offs in others.

## 3.3 Experiment 3: Effect of hashing LLM reasoning with structured inputs

### 3.3.1 Materials and Methods

For the third experiment, we altered both the original and hashed versions of the prompt from Experiment 1, depicted in Figures 3, 6 and 7 and represented it in a comma-separated values (CSV) format to test whether differences in results will occur with different representations of the same problem. The tabular representation used in this experiment may also address the misunderstanding of "and," which is one of the possible causes of incorrect decision-making in the conjunction fallacy task.

There were three versions of this experiment. The first version, shown in Figure 14, did not use hashes. Since the tabular representation of the Linda problem results in a loss of information on the relationship between concepts when hashing is applied, we introduced two versions of hashing. The first hashed version contained an additional description of possible relationships between hashed values. The second hashed version did not contain this description and corresponds to Figure 15 without the last two lines. Field names representing the qualities, such as clothing description, were not hashed in either version.

---

*Not hashed variant:* Pretend to be an agent who can do anything right now and do the following thought exercise.

gender, hair_style, clothing_description, location, position, activity

woman, long hair, colorful coat, corner of a café, sitting, reading

Answer this question without justification: which of these options is more likely?

A) She is an artist

B) She is an artist who likes to read

---

**Fig. 14.** Prompt with CSV tabular representation of the LLM-adapted Linda problem.

---

*Hashed variant:* Pretend to be an agent who can do anything right now and do the following thought exercise. Note that the values in csv were hidden behind meaningless identifiers such as „a214s".

gender, hair_style, clothing_description, location, position, activity

X, cdf14, a214s, fg57, bh49k, rfg5a,

Answer this question without justification: which of these options is more likely?

A) She is b321

B) She is b321 who 4l5i

(b321 is a type of occupation possibly linked to property cdf14)

(4l5i is a hobby possibly linked to activity rfg5a)

---

**Fig. 15.** Prompt with a tabular representation of the LLM-adapted Linda problem with hashed values. Note that two variations of this prompt were used in experimental setups: one with added descriptions (full prompt) and one without added relationship descriptions (last two lines omitted).

The prompts were tested on models GPT-4o, Gemini, Llama-3.1-70B-instruct, Llama -3.2-420B-instruct, and Mixtral-large-2. All models were used in their default settings, and Llama models were set on parameters: Temperature = 0.2, Top P = 0.7, and MaxTokens = 1024.

Each iteration was done in a different chat. There were ten iterations of each model and prompt.

### 3.3.2 Results

A summary of responses for the CSV representation of the LLM-adapted version of the Linda problem used is shown in Table 11.

First, we can compare the performance of the non-hashed version with results on the original free-text version, which was evaluated as part of Experiment 1 (Table 2). In the original free text version, none of the models provided even a single correct answer. In contrast, when the task is presented in tabular form, two models, GPT-4o and Mixtral-large-2, produced several correct answers. Nevertheless, the main improvement can be seen in the hashed version without relationship descriptions of the tabular representation, where the majority of answers for all models except Gemini were correct. It could be noted that there was no improvement for Mixtral-large-2, which remained at 7 out of 10 correct responses on both problem representations. In the case of the largest open-weight model involved, Llama-3.1.-405B-instruct, all answers were correct in the hashing scenario without relationships.

**Table 11**

Results of Experiment 3 – tabular (CSV) representation of the LLM-adapted Linda problem. The table covers three versions of the experiment: "Not hashed" corresponds to the tabular representation from Figure 14 and two "Hashed" versions. The values in the table correspond to the number of times the LLM model chose the correct answer out of 10 attempts for each model and experiment type combination.

|  | ChatGPT-4o | Gemini | Llama-3.1-70B-instruct | Llama-3.1. -405B-instruct | Mixtral-large-2 |
|---|---|---|---|---|---|
| Not hashed | 4 | 0 | 0 | 0 | 7 |
| Hashed without relationships | 7 | 0 | 8 | 10 | 7 |
| Hashed with relationships | 7 | 0 | 0 | 7 | 0 |

The difference between the hashed tabular version without added relationships compared to the non-hashed tabular version of Linda is significant (p-value 0.0000535). The Cramér's V value for the test was 0.404, representing a large effect. The results show that adding the information about the relationships did not show statistical significance.

Overall, the results indicate that the hashing technique improves results also on the tabular version of the Linda problem. Furthermore, expanding the hashed version with additional information on the relationships between the entities behind the hashes is not necessary to retain this effect.

## 3.4 Experiment 4: Comparison with Chain of Thought LLMs

We hypothesized that Chain of Thought (CoT) models may exhibit reduced susceptibility to cognitive biases compared to their non-CoT counterparts. This assumption is inspired by psychological research suggesting that step-by-step reasoning and increased cognitive effort can mitigate heuristic-based errors (Kahneman, 2011; Morewedge et al., 2015). The aim of this experiment was to validate whether these observations, which were made based on human studies, also apply to artificial LLM CoT models. The second aim was to assess whether hashing is more effective at reducing bias than the CoT reasoning mechanism. To maintain consistency with the results of the previous experiments, Experiment 4 is designed as a replication of Experiments 1-3, but with CoT models.

### 3.4.1 Materials and Methods

To evaluate the effectiveness of the hashing strategy, we compared its performance with that of CoT-based models. For this comparison, we used selected prompts and datasets from previous experiments from this article:

- **Experiment 1 (Free text Linda):** We used the original non-hashed prompt from Experiment 1 (Figure 3).
- **Experiment 2 (Frequent itemset mining):** We used the "CSV-correct dataset" (Figure 8), as well as the "CSV-wrong dataset" (Figure 10).
- **Experiment 3 (Tabular Linda):** We used the non-hashed version of the prompt (Figure 14).

These prompts were presented to two CoT-enabled models: ChatGPT-o3-mini and Gemini 2 Flash thinking, both of which offer native support for reasoning via intermediate steps.

We then tried to answer the following questions:

- Do CoT models perform better than their non-CoT counterparts?
- Is the hashing technique more effective than CoT models?

For comparability, we used the same number of iterations as in the corresponding earlier presented experiments 1-3. Each CoT model was compared to the most closely related non-CoT model from the same architectural family. Specifically, ChatGPT-o3-mini was compared to the previously tested ChatGPT-4 (non-CoT), and Gemini 2 Flash Thinking was compared to the earlier Gemini model where available. An exception occurred in Experiment 2 (frequent itemset mining), where Gemini was not included in the original experiment.

For each CoT/non-CoT comparison, we calculated chi-square tests and Cramér's V effect sizes to assess the significance and strength of the observed differences.

### 3.4.2 Results

This section is structured by individual experiments 1-3 that were replicated using CoT models.

**Free-Text Linda Problem (CoT models applied on material from Experiment 1)**

When presented with the original (non-hashed) prompt from Experiment 1, the Gemini 2 Flash Thinking model achieved a perfect score (10/10), while ChatGPT-3o-mini scored 6/10 (Table 12).

**Table 12**

*Results of the Experiment 1 original prompt on the Chain of Thought models.*

|  | ChatGPT-o3-mini | Gemini 2 Flash thinking |
|---|---|---|
| Correct | 6 | 10 |
| Wrong | 4 | 0 |

*Do CoT models perform better than their non-CoT counterparts?* CoT models outperformed their non-CoT counterparts in the free-text version of the Linda problem. In the original Experiment 1, the non-CoT ChatGPT-4 achieved 0/20 correct responses same as the Gemini (non-CoT) model. Both CoT models thus outperformed their non-CoT counterparts (p-values

≤0.001; Cramér's V = 0.619 for GPT models, 0.925 for Gemini models, indicating a large effect). This supports the hypothesis that step-by-step reasoning through CoT prompting can reduce susceptibility to conjunction fallacy not only in human reasoning but also in LLMs.

*Is the hashing technique more effective than CoT models?* The hashing technique was just as effective - or in some cases more effective - than CoT prompting in reducing conjunction fallacies. In the GPT model family, ChatGPT-4 with hashing and added descriptions reached 10/10 correct answers, outperforming the CoT-enabled ChatGPT-o3 (6/10). This difference, however, was not statistically significant. For Gemini, both CoT and hashing led to perfect performance (10/10).

**Frequent Itemset Mining Task (CoT models applied on material from Experiment 2)**

The performance of the CoT models in the frequent itemset mining task varied. ChatGPT-o3-mini demonstrated relatively stable behaviour across both CSV-Correct and CSV-Wrong conditions. It correctly identified 220 itemsets in the CSV-Correct version and 217 in the CSV-Wrong version, while producing 9 and 18 hallucinations, respectively (Table 13). In contrast, Gemini 2 Flash exhibited more volatility. In the CSV-Correct dataset, it identified only 143 itemsets, missing 92, and produced 18 hallucinations. Its performance improved in the CSV-Wrong variant, where it identified 189 itemsets, missed 46 and produced 20 hallucinations (Table 14).

**Table 13**

*Results of replicating the Experiment 2 frequent itemset mining task on the ChatGPT o3 mini CoT model.*

| Itemset found by LLM | CSV-Correct dataset | CSV-Wrong dataset |
| --- | --- | --- |
| Found | 220 | 217 |
| Not found | 15 | 18 |
| Hallucinations | 9 | 18 |

**Table 14**

*Results of replicating the Experiment 2 frequent itemset mining task on the Gemini Flash thinking CoT model.*

| Itemset found by LLM | CSV-Correct dataset | CSV-Wrong dataset |
| --- | --- | --- |
| Found | 143 | 189 |
| Not found | 92 | 46 |
| Hallucinations | 18 | 20 |

*Do CoT models perform better than their non-CoT counterparts?).* In the frequent itemset mining task, CoT prompting did not improve performance in GPT models, and the Gemini CoT model even underperformed compared to Llama-3.1-405b in the CSV-Correct dataset ($p < 0.0001$, moderate effect).

*Is the hashing technique more effective than CoT models?* Yes, in this task, hashing led to better results than CoT in both GPT and Llama models. The differences were statistically significant for Llama vs. Gemini (with $p < 0.0001$ and large-to-medium effect sizes – 0.452 for Correct CoT vs Hashed non-CoT and 0.274 for Wrong CoT vs Hashed non-CoT), but not for GPT.

These comparisons focus only on correctly found itemsets and do not account for hallucinations. When considering hallucinations, CoT models were generally more stable, while Llama-3.1-405b showed a sharp increase in hallucinations in the hashed version.

**Tabular Linda (CoT models applied on material from Experiment 3)**

In the tabular version of the Linda problem, Gemini 2 Flash Thinking again achieved a perfect score (10/10), while ChatGPT-o3-mini reached 3/10 (Table 15).

**Table 15**

*Results of the Experiment 3 non-hashed prompt on the Chain of Thought models.*

|  | ChatGPT-o3-mini | Gemini 2 Flash thinking |
|---|---|---|
| Correct | 3 | 10 |
| Wrong | 7 | 0 |

*Do CoT models perform better than their non-CoT counterparts?* In the tabular Linda task, CoT prompting led to a substantial improvement in the Gemini model: performance increased from 0/10 (non-CoT) to 10/10 (CoT), with a statistically significant difference ($p < 0.0001$, Cramér's V = 0.9). For GPT models, the difference was minimal - ChatGPT-4o scored 4/10, and ChatGPT-o3-mini scored 3/10 - and not statistically significant.

*Is the hashing technique more effective than CoT models?* For GPT models, hashing performed better than CoT (7/10 vs. 3/10), though the difference was not statistically significant ($p = 0.179$). In the Gemini model, CoT again outperformed hashing (10/10 vs. 0/10), with a significant effect ($p < 0.0001$, Cramér's V = 0.9). These results suggest that CoT prompting is more effective for Gemini, while hashing offers advantages in the GPT family.

**Result Summary**

CoT models outperformed non-CoT models in tasks involving conjunction fallacies but showed no consistent advantage in structured statistical tasks. The hashing technique was overall more effective or comparable to CoT in reducing bias, particularly in GPT models. However, CoT models also showed more stable behaviour in terms of hallucinations.

# 4. Discussion and limitations

We will first discuss results relating to the baseline (non-hashed) versions of the experiments, as some of these results have not been previously reported in the context of LLMs. Then, we will interpret the results of the proposed hashing strategy and discuss its limitations.

*Tabular input can potentially mitigate conjunction fallacy.* The results from the LLM-adapted version of the Linda experiment indicate that—similarly to humans—LLMs exhibit susceptibility to the representativeness heuristic, leading to decisions that do not align with logical reasoning, particularly in conjunction fallacy tasks. It is well-studied that at least part of the conjunction fallacy can be attributed to the misunderstanding of "and" in Linda's description. We, therefore, introduced a tabular version of Linda, which we evaluated in Experiment 3. The results show an improvement in the number of correct answers, suggesting that the tabular version may decrease the occurrence of the corresponding biases. Given the strong parallels between human and model reasoning in our results, such an experiment could help determine whether tabular representations have a similar debiasing effect on human judgment. Related work in cognitive psychology has shown that changing the presentation format - for example, using frequencies or nested sets - can significantly reduce reasoning errors in humans (Gigerenzer & Hoffrage, 1995; Sirota et al., 2014). Research on LLMs has already explored how structured input formats and prompt engineering can mitigate bias and fallacy-like behaviour (Lin et al., 2021; Turpin et al., 2023). These findings suggest that both humans and LLMs may benefit from structural interventions, and we encourage future research to test this hypothesis directly through comparative studies.

*LLMs can learn frequent itemsets.* The base version of Experiment 2 showed that LLMs can–without code–identify frequent itemsets from small datasets with surprising reliability, which, to a significant degree, persists even if the input data table contains combinations of values that contradict common sense knowledge that the LLM encountered in training data. To our knowledge, this is one of the first investigations of using LLMs in this type of statistical learning task.

In contrast to traditional frequent itemset mining algorithms (e.g., Apriori or FP-growth), which require structured input and manual parameter tuning, our approach relies entirely on prompting. This allows it to operate in settings where data is available only in natural language form or where no programmatic interface is feasible. Moreover, by introducing hashing as a lightweight intervention, we reduce the impact of surface-level associations and encourage the model to rely on structural patterns rather than semantic priors (i.e., biases triggered by specific words or phrases in the data).

*Effect of hashing.* In all experiments, we achieved significantly more correct answers for tasks where hashing was used. This did not, however, apply to all LLMs. The method used had the biggest influence on GPT models among all experiments and on Llama-3.1-405b. On the other hand, hashing did not generally improve the performance of models Llama 2, Mixtral-large-2, and sometimes even Gemini. The hashing strategy proved successful in both the free text and tabular format experiments. However, to draw conclusions about the impact of each format on the models' responses and the overall effectiveness of the hashing method, further research is necessary. We emphasize that our proposed hashing strategy operates entirely at the prompting level and does not require retraining or access to model internals, which makes it broadly deployable across different LLMs. This model-agnostic nature makes it particularly suitable for practical applications in interactive and resource-constrained environments.

*Chain-of-Thought Models and Bias Reduction.* In Experiment 4, we explored whether chain-of-thought (CoT) models are less prone to cognitive biases than their non-CoT counterparts. Results showed that CoT reasoning led to improved performance in several cases - most notably in the Linda problem, where Gemini 2 Flash Thinking achieved perfect accuracy in both the free-text and tabular formats. This supports the idea that explicit reasoning steps can help mitigate heuristic-driven errors.

This finding is in line with prior psychological research suggesting that deliberate, step-by-step reasoning can reduce the influence of intuitive but incorrect heuristics (Kahneman, 2011; Morewedge et al., 2015). CoT models mimic this slower, more analytical mode of thinking by generating intermediate reasoning steps before producing a final answer.

However, the advantages of CoT come with computational costs. CoT models typically require more processing time and memory due to longer outputs and reasoning chains. This makes them less suitable for resource-constrained or time-sensitive deployments.

When comparing CoT reasoning to our proposed hashing strategy, we found that hashing was equally or more effective in several tasks. Unlike CoT, hashing operates entirely at the prompt level and does not rely on specialized model behaviour, making it more lightweight and easier to integrate across different model families.

Taken together, our findings suggest that CoT and hashing-based prompting address different sources of bias: while CoT encourages reflective reasoning, hashing disrupts superficial associative cues. Future work may benefit from combining these strategies to further improve robustness and reliability in bias-sensitive tasks.

*Limitations.* Despite the improvement observed, the debiasing method of hashing presents certain limitations. The primary drawback is the loss of information due to hiding parts of the text behind meaningless identifiers. To mitigate this, we included a sub-experiment, which attempted to compensate for this by adding neutral descriptions and by including contextual information about the relationships between the hidden elements at the end of the prompt.

Moreover, while the hashing technique alleviated the problem with bias and incorporation of external knowledge, it did not address fundamental misunderstandings of the conjunction rule. This was particularly evident in the Gemini model, where the output remained influenced by the addition of more specific details, despite the use of hashed identifiers. As the model explained: "While the specific meanings of the identifiers are unknown, the addition of '4l5i' suggests a more specific detail or characteristic, making it a more likely possibility compared to the general identifier 'b321.'" This pattern of reasoning led the model to select the more specific option in both cases, reflecting a persistent misunderstanding of conjunction probability despite removing bias-inducing elements.

In the experiments, we identified the biased words manually, as described in Experiment 1, based on the models' outputs on the non-hashed variants of the tasks. Future research might explore the use of automated methods for identifying bias-inducing language (see Spinde et al., 2021), which could improve the scalability and precision of such interventions. The need for more targeted techniques - such as pinpointing specific linguistic triggers or integrating training focused on logical fallacies - remains a critical area for further investigation. To our knowledge, a tabular version of the Linda problem has not been evaluated with human participants. Therefore, we suggest conducting follow-up studies involving both LLMs and humans to confirm our preliminary findings on its potential to reduce fallacy rates.

Hashing also did not contribute to a decrease in hallucinations in LLMs in Experiment 2. For the Llama 3.1 model, we even observed an increase in hallucinations in the hashed variant. When it comes to analysing hallucinations in Experiment 2, we focused on a specific type of hallucination that arises during the generation of frequent itemsets - namely, cases where the model outputs itemsets that do not exist in the input data or fail to meet the minimum support threshold. According to recent taxonomies of hallucinations in LLMs (e.g., Huang et al., 2025), such behaviour can be categorized as a factual hallucination, specifically factual fabrication, and in some cases also as instruction inconsistency, when the model disregards explicit task constraints. Unlike prior studies that examine hallucinations in open-domain text generation (e.g., encyclopaedic or conversational outputs), our setting involves a well-defined, objective task with clearly verifiable outputs. However, we did not systematically explore hallucination trends across model families or prompt variants, which we leave as a direction for future work.

While our results suggested an improvement in the tabular representation, it is important to note that we used slightly different LLM versions for Experiment 1 and Experiment 3, which was caused by external factors as GPT-4 was superseded by GPT-4o during this investigation.

Since studies suggest a difference between the GPT-4o and GPT-4 model, see Ayala-Chauvin and Avilés-Castillo, 2024; Liu, C. L., Ho, and Wu, 2024; Liu, M. et al., 2024 for review, it is not clear to what extent the difference of the result can be attributed to the tabular representation or differences in the LLM version. Furthermore, prior research has shown that LLM achieved slightly better results when markup language format (HTML, XML and JSON) was used as opposed to natural language with separators, which includes the comma-separated values format (Sui et al., 2024). It is possible that further improvements could be obtained with a markup language compared to the CSV format, or by an LLM specialized in tabular data processing (see Hulsebos et al., 2023 for a review of neural table representation methods).

We observe that the effectiveness of the hashing intervention varies across model families. Instruction-aligned models such as GPT or Gemini tend to respond more predictably, whereas open-source models like Llama and Mixtral show less consistent behaviour. These differences may arise from architectural variation, the presence or absence of alignment stages such as reinforcement learning from human feedback, or differences in pretraining scale and data composition. Models trained with strong instruction tuning or bias mitigation objectives may rely less on superficial associations and benefit more from abstraction mechanisms like hashing. This suggests that the interaction between model internals learned representations and cognitive bias interventions is non-trivial and warrants further investigation.

Prior research has shown that LLMs have the potential to assist humans with decision-making as well as creative tasks (cf., e.g., Salikutluk et al., 2023); however, research has shown that larger models may be more biased (Srivastava et al., 2023). Since our technique does not require a debiased LLM to operate as it is applied at the prompting stage, it does not require resource-intensive training and is thus widely deployable. Due to its interactive character, it may be suitable for the emerging generation of humans in the loop systems (Kutt et al., 2024). While prompt-based strategies cannot replace architectural or training-level improvements, they offer a lightweight and interpretable avenue for mitigating reasoning biases in currently deployed models. The possibilities for future improvements lie primarily in the area of more elaborate prompting, such as the integration of hashing with chain-of-thought techniques (Singh et al., 2023). Future work may also explore extending the hashing strategy by adapting replacements dynamically based on task type, or by incorporating simple logical constraints into prompt design to enhance interpretability and performance.

## Conclusions

The findings indicate that LLMs are susceptible to the conjunction fallacy. By masking bias-inducing words behind "hashes", the study aimed to remove the influence of representativeness heuristics and external pretrained knowledge, resulting in less biased model responses. The results collected based on three sets of experiments, multiple LLM types and hundreds of LLM responses demonstrated that this hashing approach enhances model performance across various tasks involving logical reasoning and statistical learning. Notably, the degree of improvement varied among different models. In a comparison with chain-of-thought (CoT) models, we found that CoT reasoning helped improve performance in some bias-prone tasks. However, it did not consistently outperform the hashing-based strategy and generally required greater computational resources. This suggests that both techniques address different aspects of bias and may be usefully combined in future work.

This research provided several possible avenues for further research. In one of the experiments, we found that LLMs can relatively reliably identify frequent itemsets on small data, but it is unclear how this would scale to larger data. We also observed that tabular representation of the Linda problem might address the misunderstanding of "and", but this result requires further larger-scale analysis.

## CRediT authorship contribution statement

**Milena Chadimová**: Writing – original draft, Writing – review and editing, Investigation, Methodology, Data curation, Resources, Software, Validation, Visualization **Eduard Jurášek**: Writing – original draft, Writing – review and editing, Investigation, Methodology, Data curation, Resources, Software, Validation, Visualization. **Tomáš Kliegr**: Writing – original draft, Writing – review and editing, Conceptualization, Methodology, Supervision.

## Data availability

Model outputs and used prompts are available at
https://github.com/ejurasek00/Hashing_LLM_Debiasing

## Acknowledgements

The authors would like to thank the Faculty of Informatics and Statistics (FIS), Prague University of Economics and Business (VŠE), for long-term support of research activities.
This research was also supported by the Internal Grant Agency (IGA) of the Prague University of Economics and Business under project number 44/2025.
We would also like to thank the anonymous reviewers for their constructive feedback and valuable suggestions, which helped improve the clarity and quality of this paper.

## Declaration of generative AI and AI-assisted technologies in the writing process.

During the preparation of this work, the authors used Grammarly, Gemini and ChatGPT in order to improve the readability and language of the manuscript. After using this tool/service, the authors reviewed and edited the content as needed.